\documentclass[letterpaper, 10 pt, conference]{ieeeconf}  
  
\IEEEoverridecommandlockouts                              

\usepackage[table]{xcolor}
\usepackage{mathptmx} 
\usepackage{times} 
\usepackage{bbding}
\usepackage{graphicx}
\usepackage{float}
\usepackage{diagbox}
\usepackage{arydshln} 
\usepackage{bbm}
\usepackage{wrapfig}
\usepackage{mathrsfs}
\usepackage{listings} 
\usepackage{physics}

\usepackage{graphics} 
\usepackage{float}
\usepackage{multirow}
\usepackage{amsmath}
\usepackage{bbm}
\usepackage{amssymb}
\usepackage{mathtools}
\usepackage[utf8]{inputenc} 
\usepackage[T1]{fontenc}    
\usepackage{hyperref}       
\usepackage{cleveref}
\usepackage{url}            
\usepackage{booktabs}       
\usepackage{amsfonts}       
\usepackage{nicefrac}       
\usepackage{microtype}      
\usepackage{xcolor}         
\usepackage{algorithm}
\usepackage{algorithmicx}
\usepackage{algorithm} 
\usepackage{algpseudocode}
\usepackage{fancyvrb}
\usepackage{fvextra}
\usepackage{csquotes}
\usepackage{threeparttable}
\usepackage{cuted}%
\usepackage{afterpage}%
\usepackage{caption}
\usepackage{subcaption}
\usepackage{csquotes}
\usepackage{romannum}
\definecolor{darkred}{rgb}{0.5, 0.0, 0.0}

\newcommand{\wzl}[1]{\textcolor{black}{#1}}

\title{\LARGE \bf
RAPID Hand Prototype: Design of an Affordable, Fully-Actuated \\Biomimetic Hand for Dexterous Teleoperation
}

\author{
Zhaoliang Wan$^{1*}$, Zida Zhou$^{2*}$, Zetong Bi$^{1}$, Zehui Yang$^{1}$, Hao Ding$^{1}$, Hui Cheng$^{1\dagger}$
\thanks{* indicates equal contribution}
\thanks{$^{1}$ School of Computer Science and Engineering, Sun Yat-sen University.}
\thanks{$^{2}$ ORBOT Ltd.}
\thanks{
$\dagger$ Corresponding to chengh9@mail.sysu.edu.cn}
}

\begin{document}
 

\newcommand{\sset}{\mathcal{S}}
\newcommand{\frees}{\mathcal{S}_f}
\newcommand{\aset}{\mathcal{A}}
\newcommand{\eps}{\epsilon}
\newcommand{\init}{\rho_0}
\newcommand{\E}{\mathbb{E}}
\newcommand{\R}{\mathbb{R}}
\newcommand{\hS}{\mathbb{S}}
\newcommand{\hP}{\mathbb{P}}
\newcommand{\M}{\mathcal{M}}
\newcommand{\bs}{\vb*{bs}}
\newcommand{\N}{\mathcal{N}}
\newcommand{\G}{\mathcal{G}}
\newcommand{\C}{\mathcal{C}}
\newcommand{\Env}{\mathcal{E}}
\newcommand{\x}{\vb*{x}}

\newcommand{\obj}{o}
\newcommand{\objs}{O}
\newcommand{\pose}{p}
\newcommand{\conds}{C}
\newcommand{\poses}{P}
\newcommand{\initpose}{\pose^0}
\newcommand{\initposes}{\poses^0}
\newcommand{\goalposes}{\poses^g}
\newcommand{\goalpose}{\pose^g}
\newcommand{\cond}{c}
\newcommand{\size}{s}
\newcommand{\cate}{y}
\newcommand{\mask}{m}
\newcommand{\obs}{I_{\objs}}
\newcommand{\func}{f}
\newcommand{\dist}{p_{\func}}
\newcommand{\data}{\textit{D}_{\func}}
\newcommand{\score}{\vb*{\Phi}_{\theta}}
\newcommand{\posespace}{\mathcal{P}}
\newcommand{\condspace}{\mathcal{C}}

\newcommand\mydata[2]{$#1_{\pm#2}$}
\newcommand{\indicator}{\mathbbm{1}}
\newcommand{\loss}{\mathcal{L}}
\newcommand{\trans}{\mathcal{T}}




\def\eg{\emph{e.g}.} \def\Eg{\emph{E.g}.}
\def\ie{\emph{i.e}.} \def\Ie{\emph{I.e}.}
\def\cf{\emph{c.f}.} \def\Cf{\emph{C.f}.}
\def\etc{\emph{etc}.} \def\vs{\emph{vs}.}
\def\wrt{w.r.t. } \def\dof{d.o.f. }
\def\etal{\emph{et al}. }

\maketitle

\begin{strip}
\centering
\vspace{-50pt}
\includegraphics[width=\linewidth]{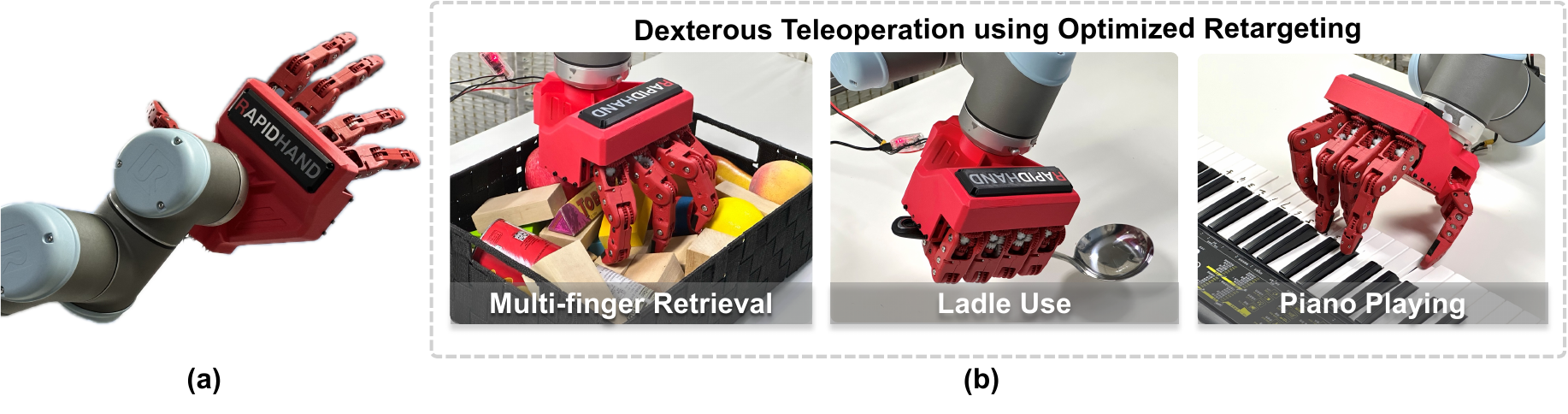}
\captionof{figure}{(a) RAPID Hand Prototype: a low-cost, fully-actuated,  five-fingered hand, each with four phalanges. (b) Dexterous teleoperation using an intuitive teleoperation interface to perform three challenging tasks: multi-finger retrieval, ladle use, and human-like piano playing.}
\label{fig: teaser}
\end{strip}

\begin{abstract}

This paper addresses the scarcity of affordable, fully-actuated five-fingered hands for dexterous teleoperation, which is crucial for collecting large-scale real-robot data within the ``Learning from Demonstrations'' paradigm. 
We introduce the prototype version of the RAPID Hand, the first low-cost, 20-degree-of-actuation (DoA) dexterous hand that integrates a novel anthropomorphic actuation and transmission scheme with an optimized motor layout and structural design to enhance dexterity. Specifically, the RAPID Hand features a universal phalangeal transmission scheme for the non-thumb fingers and an omnidirectional thumb actuation mechanism. Prioritizing affordability, the hand employs 3D-printed parts combined with custom gears for easier replacement and repair. We assess the RAPID Hand's performance through quantitative metrics and qualitative testing in a dexterous teleoperation system, which is evaluated on three challenging tasks: multi-finger retrieval, ladle handling, and human-like piano playing. The results indicate that the RAPID Hand’s fully actuated 20-DoF design holds significant promise for dexterous teleoperation.

\end{abstract}

\section{INTRODUCTION}

Dexterous manipulation \cite{ma2011dexterity, andrychowicz2020learning, chen2023sequential, chi2023diffusionpolicy, qi2023general} is a critical research area in both robotics and embodied AI. Recent advancements \cite{zhao2023learning, fu2024mobile, chi2024universal, wang2024dexcap, fang2024airexo} have been largely driven by the collection of high-quality, real-world manipulation data combined with imitation learning. A key emerging trend is the use of low-cost teleoperation systems for data collection, which offer unique advantages. These systems enable precise demonstrations with smooth trajectories, facilitating the development of policies that generalize well to new environments and tasks.

However, despite significant progress, this field faces two major challenges due to the limited availability of suitable hand hardware \cite{allegrohand, shadowhand, inspirehand, shaw2023leap, abilityhand}. First, the high cost of purchasing and maintaining multi-fingered, highly dexterous robotic hands makes them accessible only to a few research labs. Second, designing a low-cost robotic hand that mimics the full range of human hand movements remains difficult. Many affordable options have limited degrees of freedom (DoFs) or bulky fingers, limiting their usability in tasks that require fine dexterity, such as using complex tools or human-like piano playing. These limitations hinder the further development of dexterous teleoperation in robotics.

To address these issues, we introduce the prototype version of the RAPID Hand (referred to as RAPID Hand for short in this paper), an affordable, fully actuated, biomimetic robotic hand featuring five fingers, each with four phalanges (Fig. \ref{fig: teaser} (a)). It is specifically designed for dexterous teleoperation tasks to encourage broader participation in this field. Moreover, the use of off-the-shelf 3D-printed components, and custom gears ensures that parts are available to replace or repair, helping to reduce overall maintenance costs.

\begin{table*}[t!]
\caption{\textbf{Multi-finger Hands for Dexterous Manipulation Research.} Such available hands are limited and often constrained by cost and dexterity. The symbol - indicates unknown or to be determined.}
\label{tab: hand_comparsion}
\vskip 0.15in
\vspace{-4mm}
\centering
\footnotesize
\begin{tabular}{lcccccccccccc}
\toprule
\multirow{2}*{\textbf{Hands}} & \multirow{2}*{\textbf{Cost (USD)}}  & \multicolumn{3}{c}
{\textbf{Open Source}} && \multicolumn{3}{c}{\textbf{Types of Actuation}}  && \multicolumn{3}{c}{\textbf{Hand Characteristics}} \\
\cmidrule(lr){3-5} \cmidrule(lr){7-9} \cmidrule(lr){11-13}
&& Mech. & Elec. & URDF  && Tendon & Direct & Linkage && No. of Finger & DoA & DoF\\
\midrule
Trx Hand~\cite{vint6d} & -  & - & - & - && \checkmark & - & - && 3 & 8 & 8 \\
Barrett Hand~\cite{barretthand} & 25000  & - & - & \checkmark && - & \checkmark & - && 3 & 8 & 8 \\
Allegro Hand~\cite{allegrohand} & 15000  & - & - & \checkmark && - & \checkmark & - && 4 & 16 & 16\\
LEAP Hand~\cite{shaw2023leap} & 2000  & \checkmark & \checkmark & \checkmark && - & \checkmark & - && 4 & 16 & 16\\
DLR Hand \Romannum{2}~\cite{butterfass2001dlr} & $\circ$ & - & - & -  &&- & - & \checkmark && 4 & 16 & 16\\
Shadow Hand ~\cite{shadowhand} & 100000 & - & - & \checkmark  && \checkmark & - & - && 5 & 20 & 20\\
Faive Hand~\cite{toshimitsu2023getting} & - & - & - & \checkmark && \checkmark & - & - && 5 & 11 & 20\\
Inspire Hand~\cite{inspirehand} & 5000 & - & - &  \checkmark && - & - & \checkmark && 5 & 6 & 12\\
Ability Hand~\cite{abilityhand} & 20000 & - & - & \checkmark && -& - & \checkmark && 5 & 6 & 12\\
\rowcolor{violet!10} \textbf{RAPID Hand (Ours)} & 1500 & \checkmark & \checkmark & \checkmark && & & \checkmark && 5 & 20 & 20\\

\bottomrule
\end{tabular}
\vspace{-10pt}
\end{table*}

Beyond affordability, the RAPID Hand is designed to closely replicate the human hand's configuration and kinematics. The design integrates an anthropomorphic actuation and transmission scheme, with an optimized motor layout and structural configuration. Specifically, we introduce a \textbf{universal phalangeal transmission scheme} for the non-thumb fingers and an \textbf{omnidirectional thumb actuation mechanism}. This design is especially beneficial for teleoperation in scenarios where bulky fingers may obscure vision, as seen in hands like the LEAP Hand, where motors are mounted on the back of the fingers.
Besides the hand ontology design, controlling a high-DoA robotic hand presents its own challenges. To address this, we introduce a \textbf{high-DoA dexterous teleoperation interface} that bridges the embodiment gap between the human hand and the RAPID Hand, accounting for their inherent differences in geometry and kinematics.

To evaluate the RAPID Hand's performance, we conduct quantitative evaluations using two key metrics: thumb opposability \cite{chalon2010thumb} and manipulability \cite{yoshikawa1985manipulability}. Additionally, we perform qualitative evaluations,  first assessing the retargeting performance and then testing the hand across three challenging tasks—multi-finger retrieval, ladle use, and human-like piano playing—using the high-DoA teleoperation interface (Fig. \ref{fig: teaser} (b)). The results suggest that the RAPID Hand's fully actuated 20-DoF design offers promising capabilities for dexterous teleoperation.

The main contributions of this work are as follows:
\begin{itemize}
    \item A \textbf{biomimetic hardware design} that integrates a novel modular actuation and transmission mechanism within a compact structural framework, achieving four phalanges per finger and a total of 20 DoA, in a cost-effective manner.
    \item An \textbf{intuitive teleoperation interface} that optimizes retargeting of complex human hand motions to a 20-DoA, fully actuated robotic hand, enabling precise, high-DoA dexterous teleoperation.
    \item Quantitative and qualitative \textbf{experiments and analysis} demonstrate the efficacy of the proposed hand and dexterous teleoperation interface across various dexterous manipulation tasks and settings.
\end{itemize}

\section{RELATED WORK}
\subsection{Robotic Multi-fingered Hands}

We review some of the most widely used multi-fingered hands in dexterous manipulation research, as noted in Table \ref{tab: hand_comparsion}. 
The Shadow Hand \cite{shadowhand} and Allegro Hand \cite{allegrohand} are significant contributions, with the Shadow Hand offering 20 DoFs but costing around \$100,000 and requiring frequent maintenance due to its complex tendon-driven mechanism. The Allegro Hand, priced at \$15,000, has 16 DoFs with direct-drive joints, but its closed-source nature makes repairs and replacements more challenging.
For three-fingered hands, the Trx Hand \cite{vint6d} and Barrett Hand \cite{barretthand} provide 8 DoFs. The Trx Hand, developed for internal use, is not commercially available, while the Barrett Hand costs \$25,000. 
Among four-fingered hands, the LEAP Hand \cite{shaw2023leap} offers a low-cost, open-source solution. However, its motor-mounted design results in bulky finger appearance and unnatural movement, which may hinder dexterous teleoperation. The DLR Hand \Romannum{2} \cite{butterfass2001dlr} is another internal-use system that is not commercially available.
For five-fingered hands, the Inspire Hand \cite{inspirehand} and Ability Hand \cite{abilityhand} are more affordable but have underactuated joints that limit dexterity. The Faive Hand \cite{toshimitsu2023getting}, with 11 DoFs and a rolling contact joint design, struggles with proprioception accuracy due to reliance on extended Kalman filters (EKFs). Despite their advancements, these hands still face challenges such as high costs, maintenance needs, closed-source limitations, and bulkiness, which impact their performance in dexterous manipulation tasks.

\subsection{Dexterous Teleoperation Systems}

Robotic teleoperation systems are gaining attention for their potential in collecting real-world manipulation demonstrations for robotic foundation models \cite{brohan2022rt, pmlr-v229-zitkovich23a, RTX}. Various approaches have been proposed, including VR devices \cite{ding2024bunny, lin2024learning, cerulo2017teleoperation}, wearable gloves \cite{liu2019high, wang2024dexcap}, RGB cameras \cite{qin2023anyteleop, fu2024humanplus, sivakumar2022robotic}, and leader-follower hardware platforms \cite{zhao2023learning, fu2024mobile, kim2024surgical, xu2025robopanoptes}. Many systems use parallel grippers or underactuated five-fingered hands, such as the 6-DoF Inspire Hand and Ability Hand, which are cost-effective for tasks like object pick-and-place and simple tool use. However, their limited DoA and underactuation hinder performance in more complex tasks, such as pen spinning \cite{wang2024lessons}, lid twisting \cite{lin2024twisting}, and object-in-hand pose adjustment \cite{yin2024learning}. Existing retargeting methods are often optimized for low-DoA, underactuated hands. Systems like \cite{wang2024dexcap, shawleap, shaw2023leap}, using the 16-DoA LEAP Hand, still face challenges due to the embodied gap caused by differences in kinematic structures and geometries, which affect contact dynamics and complicate transferring human motions to robots. High-DoA dexterous teleoperation presents significant challenges, as it involves complex motion requiring precise control. Any suboptimal movement can lead to failures, highlighting the need for high-DoA retargeting methods to improve performance and precision in more complex tasks.

\begin{figure*}[t!]
\begin{center}
\includegraphics[width=\linewidth]{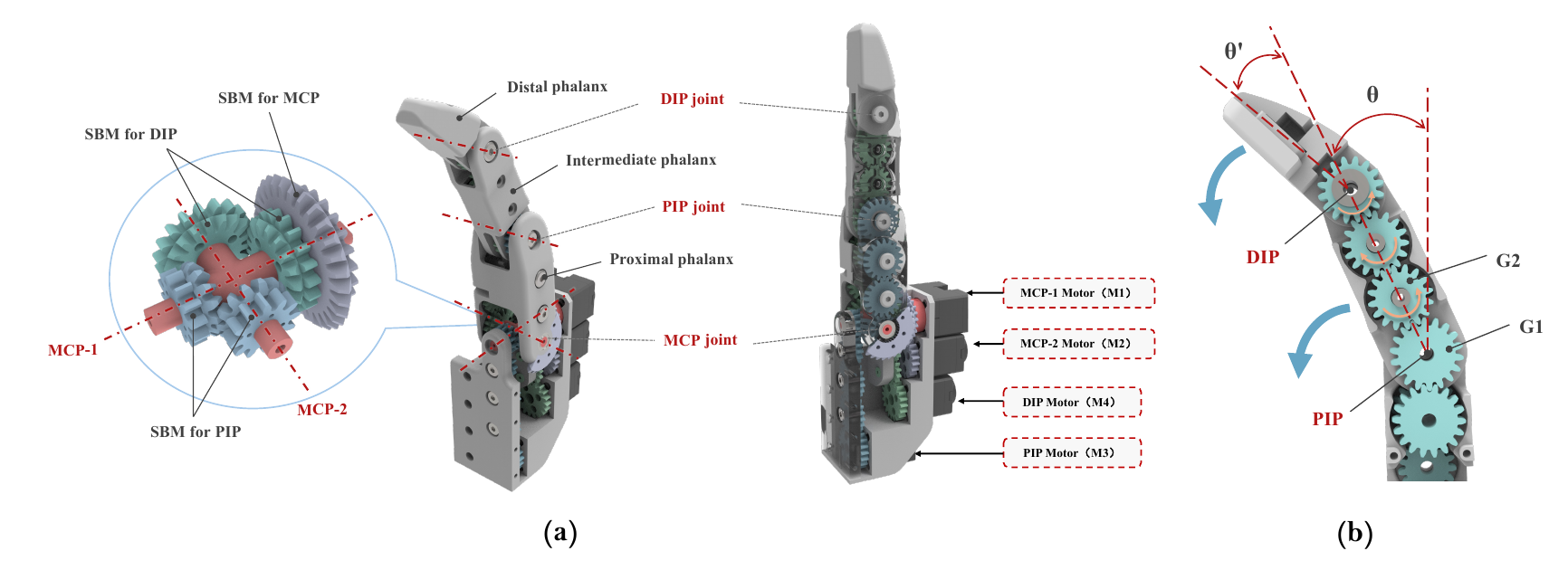}
\end{center}
\vspace{-10pt}
\caption{(a) \textbf{Multi-view of the index finger with universal phalangeal transmission scheme}. Red indicates MCP-1 transmission, purple for MCP-2, blue for PIP, and green for DIP. (b) \textbf{Effect of Joint Motion on Gear Transmission}. As the PIP motor rotates the PIP by $\theta$ and DIP motor remains stationary, gear G1 stays fixed, while G2 rotates, causing the DIP joint to rotate by $\theta'$ relative to the IP.}
\label{fig: index_finger_design}
\vspace{-15pt}
\end{figure*}

\section{DESIGN FOR DEXTERITY}
\label{sec:hand design}

The RAPID Hand is designed for dexterous teleoperation tasks to encourage wider participation in this research area. Its design adheres to the following principles:
\begin{itemize}
    \item \textbf{Affordable.} The design is cost-effective, allowing researchers to build, repair, or replace the motors by themselves after extensive use.
    \item \textbf{Human-like Dexterous.} To handle daily objects and tools, the RAPID Hand is designed to mimic the human hand's configuration, kinematics, and dexterity. 
\end{itemize}

The following sections detail how we achieve these goals.

\subsection{Simplified Hand Kinematics}

\begin{figure}[t!]
\begin{center}
\includegraphics[width=0.7\linewidth]{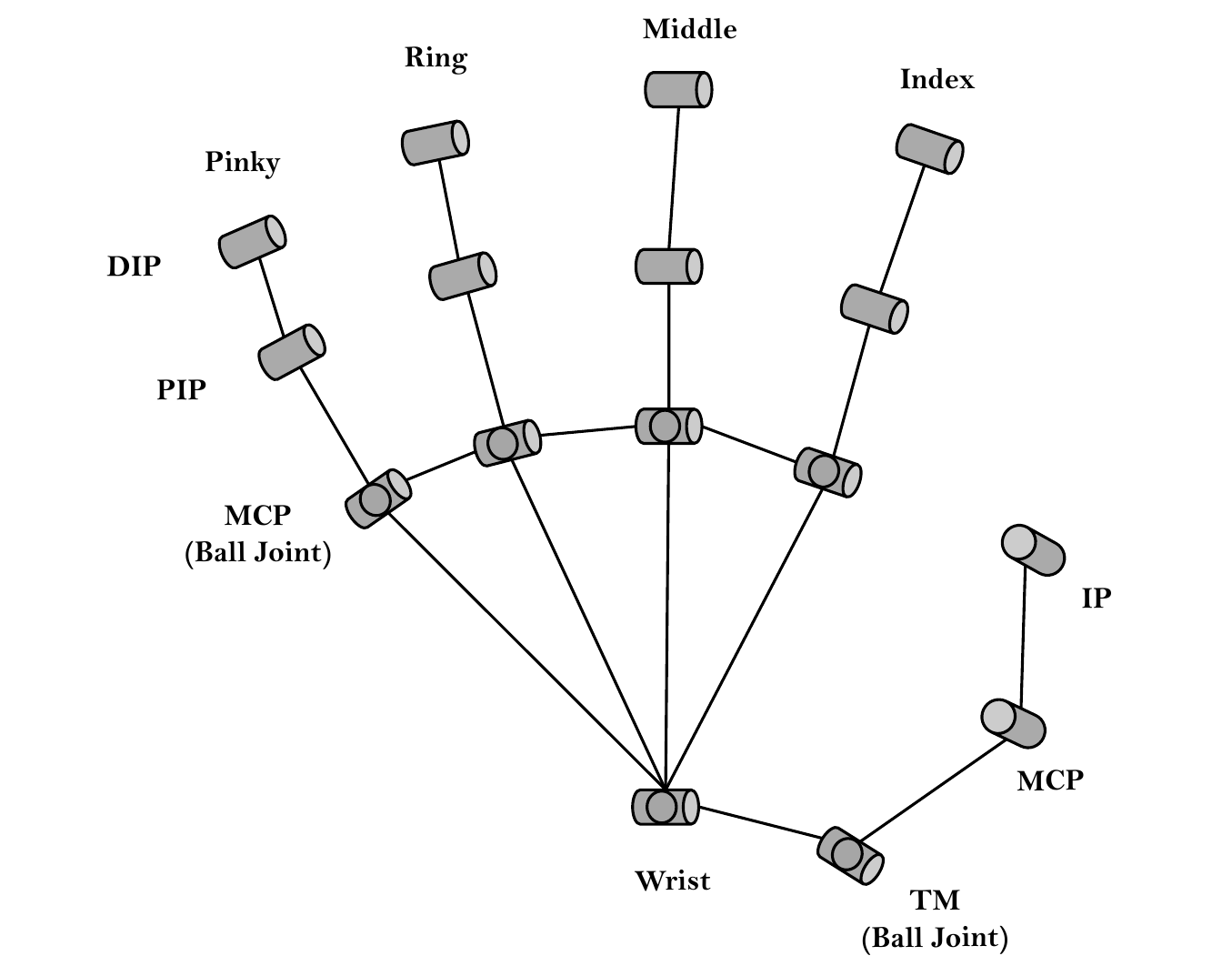}
\end{center}
\vspace{-10pt}
\caption{\textbf{Simplified human hand kinematics.} \cite{cerulo2017teleoperation} }
\label{fig: hand_kinematics}
\vspace{-15pt}
\end{figure}

\begin{figure*}[t!]
\begin{center}
\includegraphics[width=\linewidth]{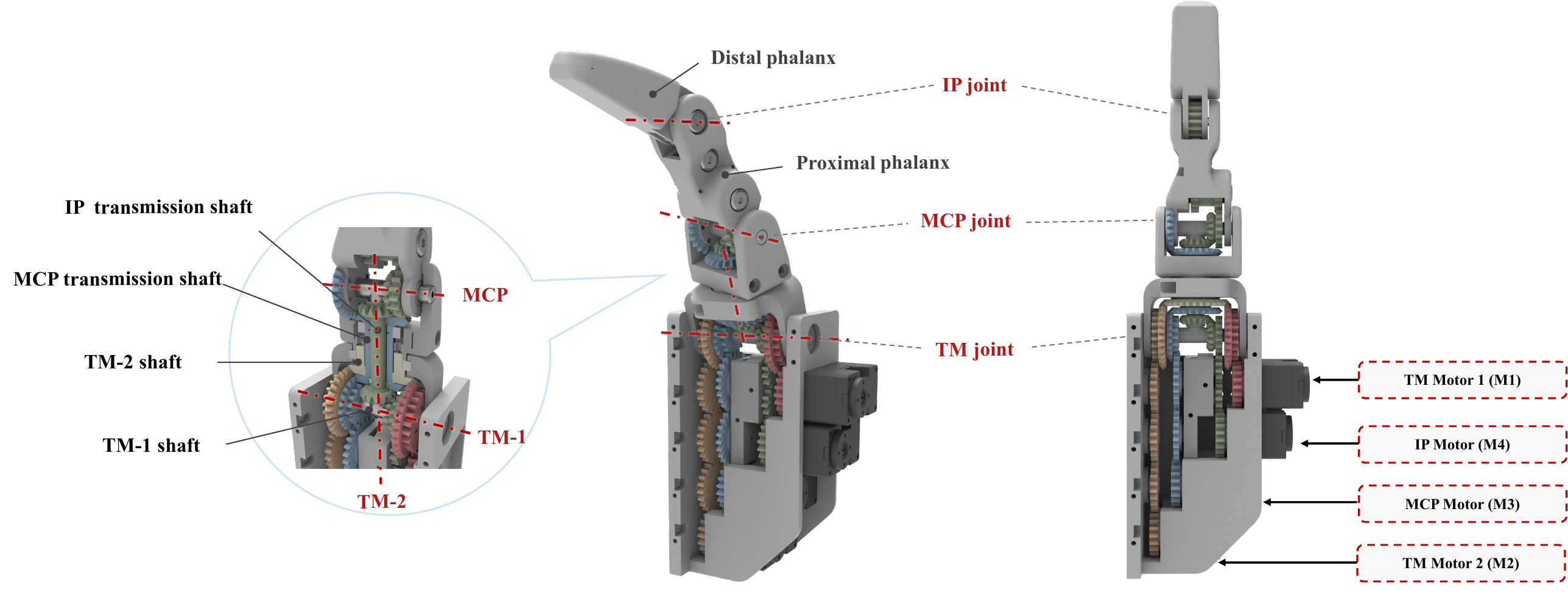}
\end{center}
\caption{\textbf{Mechanical design of the thumb finger}. The red and orange sections represent the TM joint transmission gear set, the blue section represents the MCP transmission gear set, and the green section represents the IP transmission gear set.}
\label{fig: thumb_joint}
\vspace{-15pt}
\end{figure*}

Given that most household objects and tools are designed for human hands, the RAPID Hand is designed to closely replicate human hand kinematics. As shown in Fig. \ref{fig: hand_kinematics}, the human hand comprises five fingers, each with three joints. The thumb includes an interphalangeal (IP), metacarpophalangeal (MCP), and trapezoid-metacarpal (TM) joints, while the other fingers feature distal interphalangeal (DIP), proximal interphalangeal (PIP), and MCP joints. Notably, the MCP and TM joints are ball joints, whereas the remaining are hinge joints. In the RAPID Hand, we use two separate DoFs to approximate the motion of the MCP and TM ball joints, resulting in a total of 20 DoFs. Replicating these DoFs in an affordable robotic hand poses a significant design challenge.

\subsection{Universal Phalangeal Transmission Scheme}
\label{subsec: index finger design}

For the non-thumb fingers, each features an MCP joint with two DoFs for abduction/adduction (MCP-1) and flexion/extension (MCP-2), as well as PIP and DIP joints, each with one DoF, all controlled by four servo motors $M$. 
 In some existing designs, the MCP joint (a ball joint) is approximated using two closely arranged motors \cite{lee2016kitech}, as seen in designs like the Allegro Hand and LEAP Hand. However, these designs place motors inside the fingers, making them bulky and heavy.

The RAPID Hand addresses this by introducing a \textbf{universal phalangeal transmission scheme} for the index, middle, ring, and pinky fingers. This standardized design relocates the motors to the palm to minimize finger size and weight, utilizing a gear-driven mechanism to transfer motor output into smooth, precise, coordinated finger movements, as shown in Fig. \ref{fig: index_finger_design}.

For the MCP joint, MCP-1 motor (M1) controls abduction and adduction through a cross shaft, while MCP-2 motor (M2) controls flexion and extension. M1 is directly linked to the cross shaft to drive the MCP-1 joint, whereas M2 needs to turn the bevel gear mounted on the proximal phalanx through a gear set to drive the MCP-2 joint. The transmission between motor M3 and the PIP joint is also achieved through a gear set; the initial gear of this set is connected to M3, and the terminal gear is located at the PIP joint and fixed with the intermediate phalanx. Similarly, the M4 motor connects to a gear set that drives the DIP joint through a gear attached to the distal phalanx.

One of the main difficulties with this design is aligning the motor output shafts with the MCP-1 axis, while ensuring that the rotational axes of the MCP-2, PIP, and DIP joints are perpendicular to it. To solve this issue, we have developed a component called the spur gear-bevel gear module (SBM). This component combines a spur and bevel gear into a single unit and allows for a 90-degree rotation of the transmission direction by being hinged on the cross shaft. Consequently, this solution guarantees precise motion transmission across all joints, overcoming the issues posed by the differing orientations of the rotational axes.

It is crucial to consider the transmission phenomenon illustrated in Fig. \ref{fig: index_finger_design} (b). The green gears in the figures represent the transmission gears of the DIP joint. The rotation axes of the intermediate phalanx (IP) and G1 are situated on the PIP joint, and G2 is hinged on the IP. Therefore, G1, G2, and IP form a planetary gear structure. Movement in the PIP joint leads to G2 rotating in relation to the IP, causing unintended movement in the DIP joint, even if the DIP motor (M4) remains inactive. To ensure independent control of each joint, it is essential to carry out gear transmission analysis and gear ratio calculations to establish the relationship between the output of motors and the angular motion of joints, as shown in \ref{Tab: non-thumb Kinematic Matrix}.

\begin{table}[t!]
\centering
\caption{Motor-to-Joint Kinematics for Non-Thumb Fingers}
\label{Tab: non-thumb Kinematic Matrix}
\vskip 0.15in
\vspace{-4mm}
\centering
\footnotesize
\begin{tabular}{ccccccc}
\toprule
 \multirow{1}{*}{\textbf{ }} & \multicolumn{1}{c}{\textbf{$MCP-1$}} & \multicolumn{1}{c}{\textbf{$MCP-2$}} & \multicolumn{1}{c}{\textbf{$PIP$}} & \multicolumn{1}{c}{\textbf{$DIP$}}\\
\midrule
$M_1 = \theta_1$ & $-\theta_1$ & $-\theta_1$ & $-\frac{10}{3} \theta_1$ & $-\frac{13}{12} \theta_1$ \\
\midrule
$M_2 = \theta_2$ & $0$ & $-\frac{10}{3} \theta_2$ & $-\frac{50}{33} \theta_2$ & $-\frac{155}{132} \theta_2$ \\
\midrule
$M_3 = \theta_3$ & $0$ & $0$ & $\frac{10}{9} \theta_3$ & $\frac{85}{36} \theta_3$ \\
\midrule
$M_4 = \theta_4$ & $0$ & $0$ & $0$ & $\frac{5}{4} \theta_4$ \\
\bottomrule
\end{tabular}
\vspace{-10pt}
\end{table}

\subsection{Omnidirectional Thumb Actuation Mechanism}
\label{subsec: thumb finger design}

\begin{figure}[t!]
\begin{center}
\includegraphics[width=\linewidth]{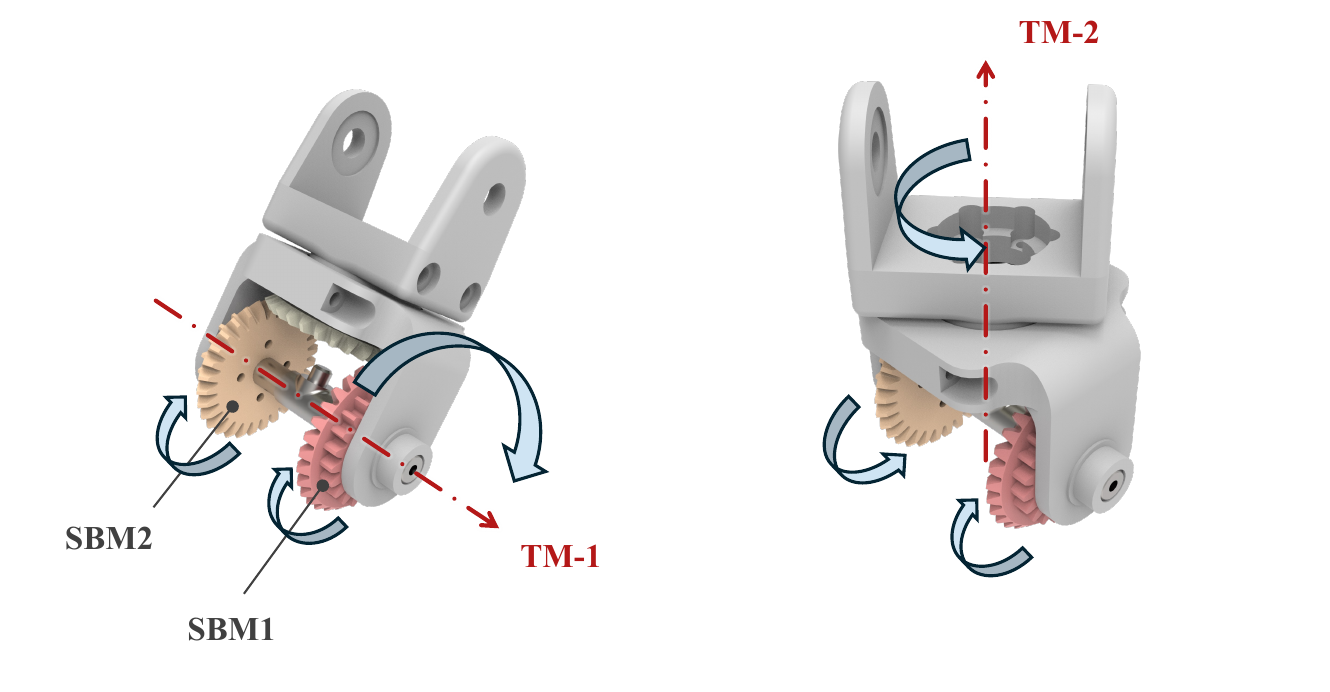}
\end{center}
\vspace{-10pt}
\caption{\textbf{Thumb TM joint actuation.} When SBM1 and SBM2 rotate at equal speeds in the same direction, the TM joint rotates around the TM-1 axis. When SBM1 and SBM2 rotate at equal speeds in opposite directions, the TM joint rotates around the TM-2 axis.}
\label{fig: thumb_mcp_driven}
\vspace{-15pt}
\end{figure}

The thumb is designed separately from the other fingers, with unique actuation requirements to support its complex movements. The actuation scheme is driven by the need for precise, omnidirectional motion to facilitate dexterous manipulation. Unlike the other fingers, the thumb requires multi-axis control to replicate human thumb movements, including flexion/extension, abduction/adduction, and rotation. To achieve this, we introduce a novel \textbf{omnidirectional thumb actuation mechanism} that integrates a differential control system for the TM joint and an efficient transmission mechanism for the MCP and IP joints.

In the omnidirectional thumb actuation mechanism, as shown in Fig. \ref{fig: thumb_mcp_driven}, the TM joint provides two degrees of freedom: flexion/extension (TM-1) and abduction/adduction (TM-2). These are powered by two motors, M1 and M2, operating through a differential mechanism that allows for independent and simultaneous control of the motions. M1 and M2 each drive a spur gear-bevel gear module (SBM1 and SBM2), which engages with the gear fixed on the TM-2 shaft. When SBM1 and SBM2 rotate at the same speed in the same direction, flexion/extension is achieved, while opposing rotations at the same speed generate abduction/adduction.The compound motion of the two DoFs can be completed by superimposing the rotational speed. This coordinated system enables smooth omnidirectional thumb movement, crucial for in-hand manipulation tasks.

Meanwhile, the MCP and IP joints present a challenge in transmission as the gear axes of their gear sets must shift from the TM-1 direction to the TM-2 direction, and then to the MCP/IP direction. To address this, a specific design is implemented at the TM joint, as shown in Fig. \ref{fig: thumb_joint}. In the case of the MCP joint, a transmission shaft with bevel gears at both ends is embedded in the TM joint. The bevel gear at one end engages with an SBM hinged on the TM-1 shaft, connected to the M3 motor via a gear set. The bevel gear at the other end engages with a bevel gear mounted on the Proximal phalanx, which is coaxial with MCP. Similarly, the transmission of the IP joint inside the TM joint is accomplished through the IP transmission shaft. The IP transmission shaft is nested inside the MCP transmission shaft, while the MCP transmission shaft is nested inside the TM-2 shaft, and the three shafts are coaxial. The motor-to-joint kinematics is shown in \ref{Tab: Thumb Kinematic Matrix}.

\begin{table}[t!]
\centering
\caption{Thumb Finger Motor-to-Joint Kinematics}
\label{Tab: Thumb Kinematic Matrix}
\vskip 0.15in
\vspace{-4mm}
\centering
\footnotesize
\begin{tabular}{ccccccc}
\toprule
 \multirow{1}{*}{\textbf{ }} & \multicolumn{1}{c}{\textbf{$TM-1$}} & \multicolumn{1}{c}{\textbf{$TM-2$}} & \multicolumn{1}{c}{\textbf{$MCP$}} & \multicolumn{1}{c}{\textbf{$IP$}}\\
\midrule
$ (M_1-M_2)/2=\theta_1$ & $\frac{10}{11} \theta_1$ & $0$ & $0$ & $0$ \\
\midrule
$(M_1+M_2)/2=\theta_2$ & $0$ & $-\frac{10}{11} \theta_2$ & $\frac{10}{11} \theta_2$ & $\frac{10}{11} \theta_2$ \\
\midrule
$M_3 =\theta_3$ & $0$ & $0$ & $\frac{5}{4} \theta_3$ & $\frac{5}{2} \theta_3$ \\
\midrule
$M_4=\theta_4$ & $0$ & $0$ & $0$ & $-\frac{5}{4} \theta_4$ \\
\bottomrule
\end{tabular}
\vspace{-10pt}
\end{table}

\subsection{High-DoA Dexterous Teleoperation Interface}
\begin{figure}[t!]
\begin{center}
\includegraphics[width=\linewidth]{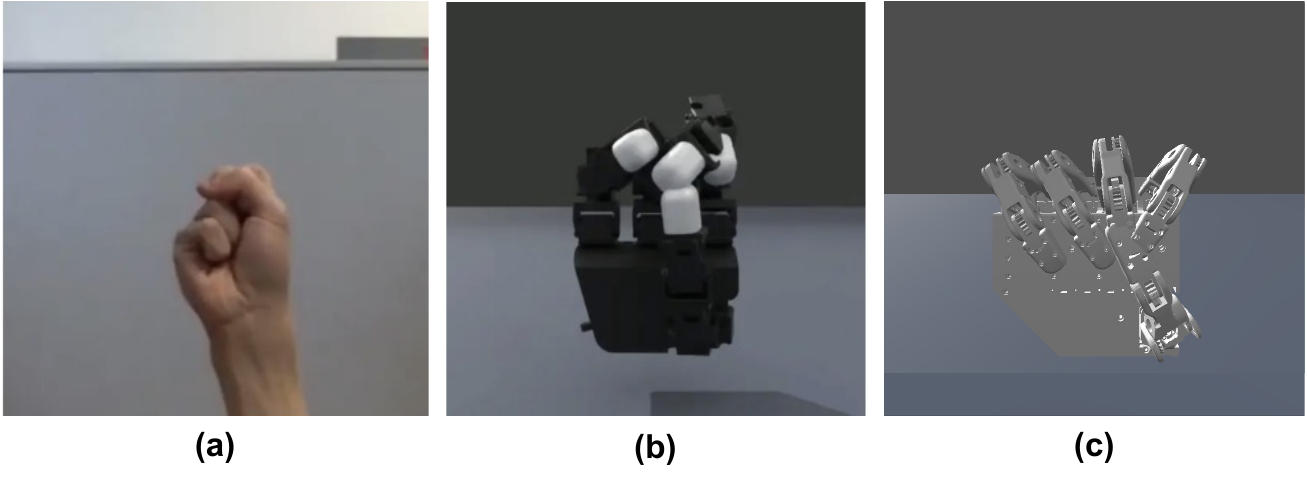}
\end{center}
\vspace{-10pt}
\caption{\textbf{Hand Retargeting Comparison}. (a) Human fist closure; (b) LEAP Hand with collisions and unnatural finger poses using \cite{qin2023anyteleop}; (c) Retargeted RAPID Hand with natural configuration.}
\label{fig: retargeting_1}
\vspace{-10pt}
\end{figure}

Despite the RAPID Hand hardware being designed to closely mimic human hand kinematics—with five fingers and four phalanges per finger—dexterous manipulation via teleoperation remains challenging due to inherent differences in kinematic structures and geometries, especially in terms of hand size and thumb motion. 
Existing methods \cite{qin2023anyteleop, ding2024bunny, cerulo2017teleoperation}, which are often optimized for underactuated or low-DoA robotic hands, tend to underperform when applied to high-DoA robotic hands. 
As shown in Fig. \ref{fig: retargeting_1}, direct mapping to the LEAP Hand often results in collisions and unnatural finger configurations, especially during fist closure.
Here, we introduce a user-friendly teleoperation interface specifically tailored for high-DoA robotic hands. This interface supports various human motion capture technologies, including cameras, VR devices, and MoCap systems. 
The key idea is to bridge the embodiment gap in retargeting by focusing on the differences in hand size and motion.

More specifically, we first refine the retargeting process to better accommodate high DoA capabilities, effectively reducing the size gap between different teleoperators with a single-shot calibration. We considered four key points on each finger, organizing the human hand's key points from the wrist to the fingertip in order:
\begin{equation}
    V_i = \{w_{i,0} \text{(wrist)}, w_{i,1}, \ldots, w_{i,n_i}\}, \quad i = 0, \ldots, 4,
\end{equation}
where \( w_{i,j} \in \mathbb{R}^3 \) denotes the raw sensor data, which shows the location of the \( j \)-th keypoint on the \( i \)-th finger relative to the wrist coordinates. For the phalange \( j \) of finger \( i \) (connecting \( w_{i,j} \) and \( w_{i,j+1} \)), we can refine the scaling ratio:
\begin{equation}
    r_{i,j} = \frac{\|f_{robot}(c_{i,j+1}) - f_{robot}(c_{i,j})\|}{\|w^*_{i,j+1} - w^*_{i,j}\|},
\end{equation}
where \( c_{i,j} \) represents predefined keypoints on the robotic hand, \( w^*_{i,j} \) indicates the positions of the human hand's keypoints obtained through a prior calibration process, and \( r_{i,j} \) is precomputed and fixed, remaining unaffected by subsequent remapping calculations.

Our goal is to measure the lengths of each joint in the human hand using different devices and compare them with those of a robotic hand. This process is simple, as it only necessitates the prior collection of your human hand data. Given that the human hand poses captured by sensors during operation can often be inaccurate and unstable, we store a more precise set of data to correct this variable. Following calibration, the coordinates \( v_{i,j} \) are derived recursively:
\begin{equation}
    v_{i,j} = \begin{cases} 
w_{i,0}, & j = 0 \\
v_{i,j-1} + r_{i,j-1}(w_{i,j} - w_{i,j-1}), & j \geq 1
\end{cases},
\label{eq:gesture_keypoint_correction}
\end{equation}
Such one-shot calibration ensures that each phalange of the human hand aligns in length with the corresponding segment of the robotic hand. This improves spatial alignment and helps accurately compute the robotic hand’s joint angles, calculated as follows:
\begin{equation}
    \min_{Q(t)} \sum_{(i,j) \in K} \|v_{i,j}(t) - f_{robot}(c_{i,j}; Q(t))\|^2 + \beta \|Q(t) - Q(t-1)\|^2,
\end{equation}
where \( K \) represents a set of user-specified keypoints, \( K \subseteq \{(i, j) | 0 \leq i \leq 4, 0 \leq j \leq n_i\} \), \( v_{i,j}(t) \) denotes the 3D coordinates of the \( j \)-th joint of the \( i \)-th finger after calibration at time \( t \), \( Q(t) \in \mathbb{R}^n \) is the vector of joint angles at time \( t \), \( f_{robot}(c_{i,j}; Q(t)) \) represents the coordinates of the robotic hand's keypoints calculated based on \( Q(t) \), and \( \beta \) is a weighting factor for joint motion smoothness. 

We utilize Sequential Least-Squares Quadratic Programming (SLSQP) \cite{boggs1995sequential, johnson2021nlopt} to solve this optimization in real-time, aiming for smooth and stable motion retargeting in dexterous teleoperation. This method effectively computes the robotic hand’s joint angles, aligning its keypoints with those of the human hand in the wrist coordinate system. Our calibration technique streamlines the process, requiring only the initial recording of an accurate and stable human hand pose without the need for subsequent parameter adjustments.

\begin{figure*}[t!]
\begin{center}
\includegraphics[width=\linewidth]{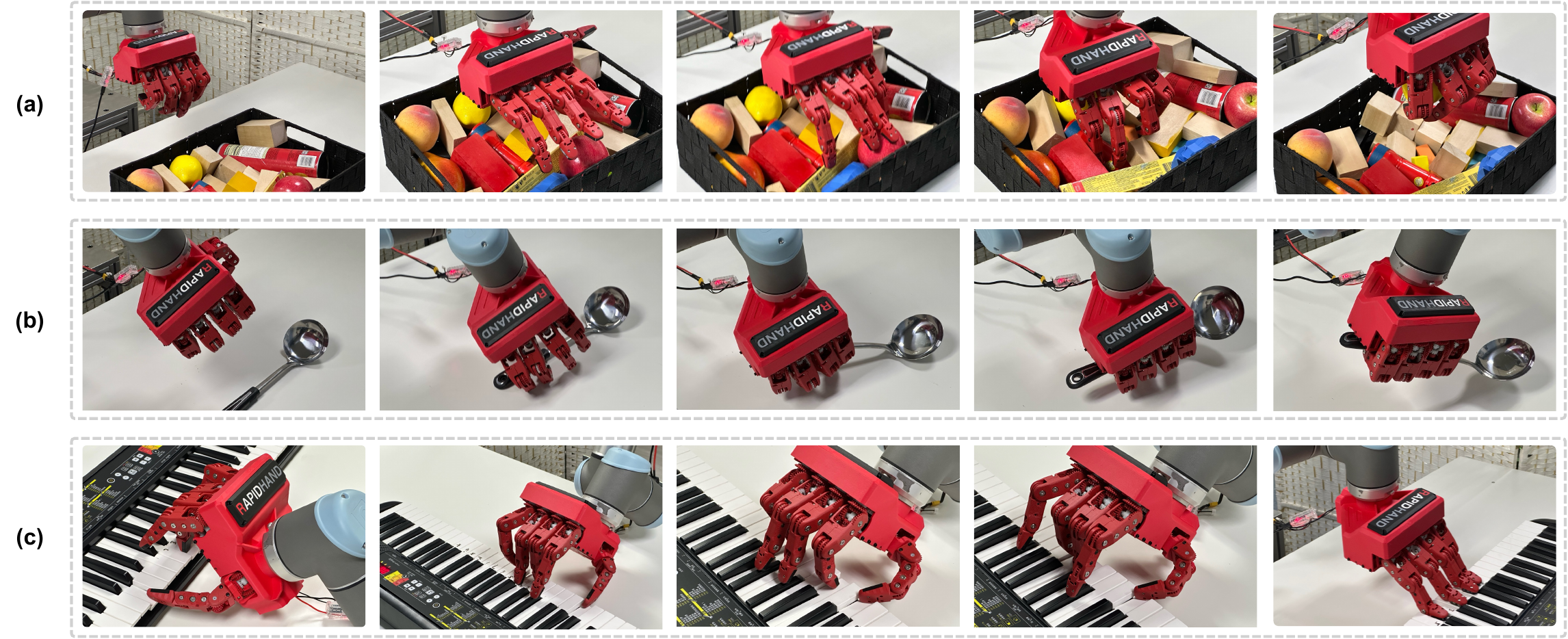}
\end{center}
\caption{\textbf{Illustrations of the dexterous teleoperation tasks.} (a) The robot performs a retrieval task, using multi-fingered non-prehensile manipulation to retrieve a target object from a densely stacked drawer. (b) The robot picks up a ladle from the table and adjusts its grip to use the ladle in a human-like manner. (c) The robot manipulates a piano with the RAPID Hand, mimicking the actions of a pianist.}
\label{fig: tasks}
\vspace{-15pt}
\end{figure*}

\section{EXPERIMENTAL VALIDATION}

In the experiments, we validate the dexterity of the RAPID Hand hardware through quantitative analysis of its manipulability metrics and qualitative evaluation of its retargeting capabilities in complex dexterous teleoperation tasks and settings.

\subsection{Quantitative Evaluation}
\label{subsec: mani. analy.}

\begin{figure}[t!]
\begin{center}
\includegraphics[width=\linewidth]{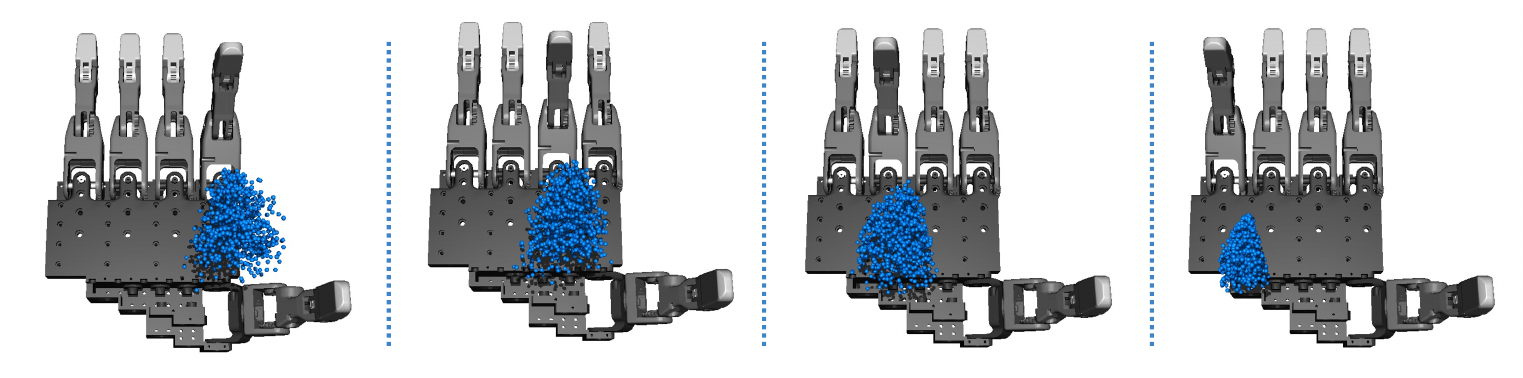}
\end{center}
\vspace{-10pt}
\caption{\textbf{Visualization of thumb opposability volume of the RAPID Hand.} Blue points represent the possible positions within the finger-to-thumb opposability of the RAPID Hand.}
\label{fig: oppo}
\vspace{-10pt}
\end{figure}

\textbf{Thumb opposability} \cite{chalon2010thumb} measures the thumb's ability to oppose other fingers, indicating the spatial range for in-hand manipulation. A more dexterous hand has a higher thumb opposability volume. We compare the thumb opposability of the Allegro Hand, LEAP Hand, and RAPID Hand, which have similar hand sizes. The LEAP Hand shows the highest opposability due to its MCP-side joint placement on the intermediate phalanx. However, this design results in a bulkier appearance and unnatural kinematics, leading to significant movement differences when retargeting. In contrast, the RAPID Hand improves on the Allegro Hand by optimizing finger design and motor arrangement, achieving a better balance between dexterity and natural motion. The additional pinky finger on the RAPID Hand further enhances opposability, as shown in the pinky-to-thumb opposability volume. Fig. \ref{fig: oppo} visualizes the RAPID Hand’s thumb opposability, with blue points representing positions within the finger-to-thumb workspace. These results suggest that the RAPID Hand’s thumb opposability benefits from its optimized mechanical transmission and motor configuration.

\begin{table}[t!]
\centering
\caption{\textbf{Finger-to-Thumb Opposability Volume} (mm$^3$)}
\label{Tab: oppo volu}
\vskip 0.15in
\vspace{-4mm}
\centering
\footnotesize
\begin{tabular}{ccccccc}
\toprule
 \multirow{1}{*}{\textbf{Hands}} & \multicolumn{1}{c}{\textbf{Index}} & \multicolumn{1}{c}{\textbf{Middle}} & \multicolumn{1}{c}{\textbf{Ring}} & \multicolumn{1}{c}{\textbf{Pinky}}\\
\midrule
Allegro Hand & 323421 & 262583 & 106047 & - \\
\midrule
LEAP Hand & 985629 & 852521 & 497977 & - \\
\midrule
RAPID Hand (Ours)& 229705 & 227154 & 173029 & 65861\\
\bottomrule
\end{tabular}
\end{table}

\begin{table}[t!]
\centering
\caption{\textbf{Manipulability Ellipsoid Volume} (mm$^3$)}
\label{tab: ellipsoid volume}
\vskip 0.15in
\vspace{-4mm}
\centering
\begin{tabular}{lccc}
\toprule
\textbf{Robot/Position} & \textbf{Down} & \textbf{Up} & \textbf{Curled} \\
\midrule
\textit{Allegro Hand} & & & \\
Linear & \wzl{246} & \wzl{48.3} & \wzl{$ 2.21 \times 10^{5}$}\\
Angular & 0 & 0 & 0 \\
\midrule
\textit{LEAP Hand} & & & \\
Linear & \wzl{$ 3.03 \times 10^{3}$} & \wzl{$ 3.03 \times 10^{3}$}  & \wzl{$ 1.36 \times 10^{5}$}  \\
Angular & \wzl{$ 1.18 \times 10^{3}$}  & \wzl{$ 5.23 \times 10^{5}$}  & \wzl{$ 2.50 \times 10^{5}$}  \\
\midrule
\textit{RAPID Hand (Ours)} & & & \\
Linear & $1.69 \times 10^{4}$ & $2.37 \times 10^{2}$ & $1.46 \times 10^{5}$ \\
Angular & $4.61$ & $4.61 $ & $4.61 $ \\

\bottomrule
\end{tabular}
\end{table}

\textbf{Manipulability} \cite{yoshikawa1985manipulability} measures a finger's dexterity in specific poses. Here, we calculate the manipulability ellipsoid volumes of the index finger at three common poses—down, up, and curled—and compare them with those of the LEAP and Allegro Hands using the 
hand's Jacobian.
As shown in Table \ref{tab: ellipsoid volume}, the RAPID Hand exhibits improved manipulability in most tested poses compared to the LEAP and Allegro Hands, reflecting its optimized ontology design for dexterous tasks.

\subsection{Qualitative Evaluation}
\label{subsec: teleoperate}

\subsubsection{Retargeting Performance}

\begin{figure}[t!]
\begin{center}
\includegraphics[width=\linewidth]{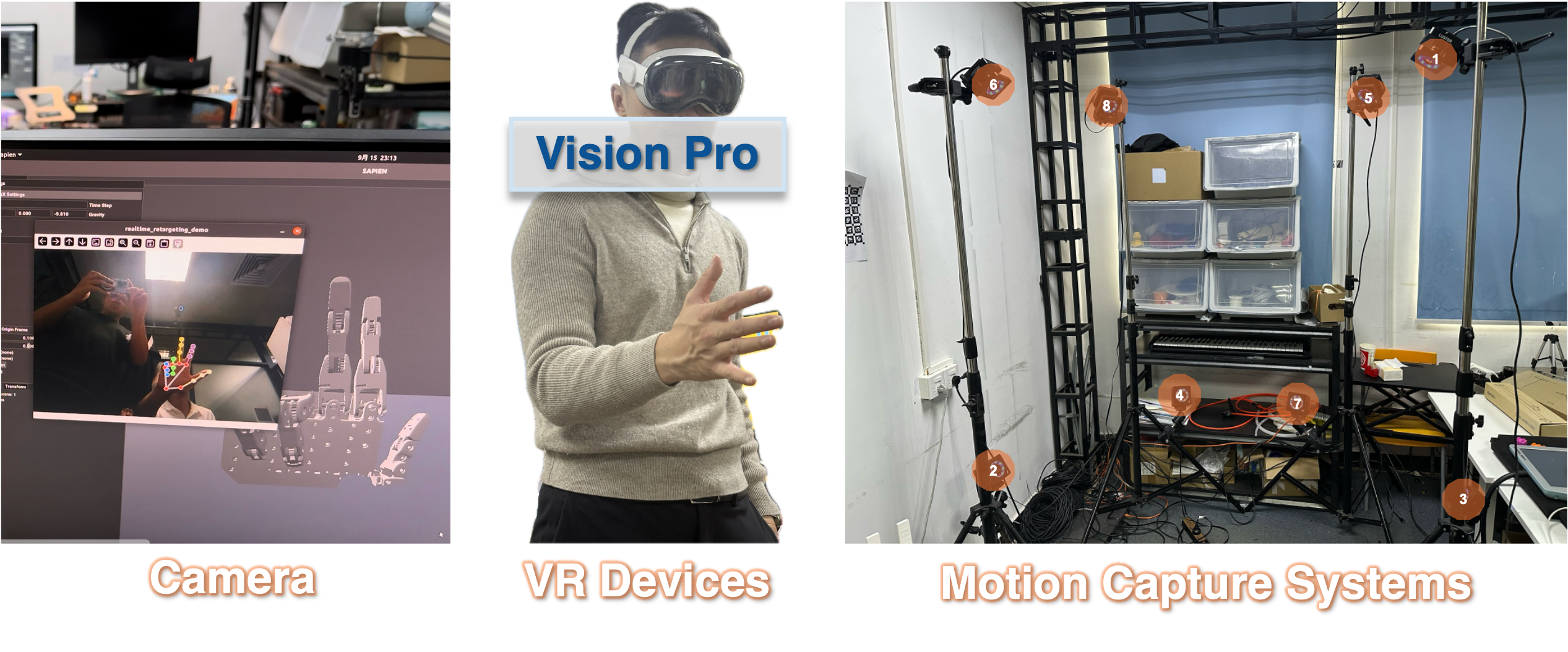}
\end{center}
\vspace{-10pt}
\caption{\textbf{Human hand poses are captured using various systems}. }
\label{fig: retargeting_3}
\vspace{-10pt}
\end{figure}

We conduct qualitative teleoperation experiments to evaluate the RAPID Hand's dexterity and showcase the capabilities of its 20-DoF fully actuated design. Retargeting performance is first assessed by visualizing the RAPID Hand’s movements in SAPIEN \cite{Xiang_2020_SAPIEN}. Human hand poses are captured using various systems—camera (Kinect Azure), AR (Apple Vision Pro), and MoCap (FZmotion)—and retargeted to the RAPID Hand via the high-DoA dexterous teleoperation interface, as shown in Fig. \ref{fig: retargeting_3}.

\begin{figure}[t!]
\begin{center}
\includegraphics[width=\linewidth]{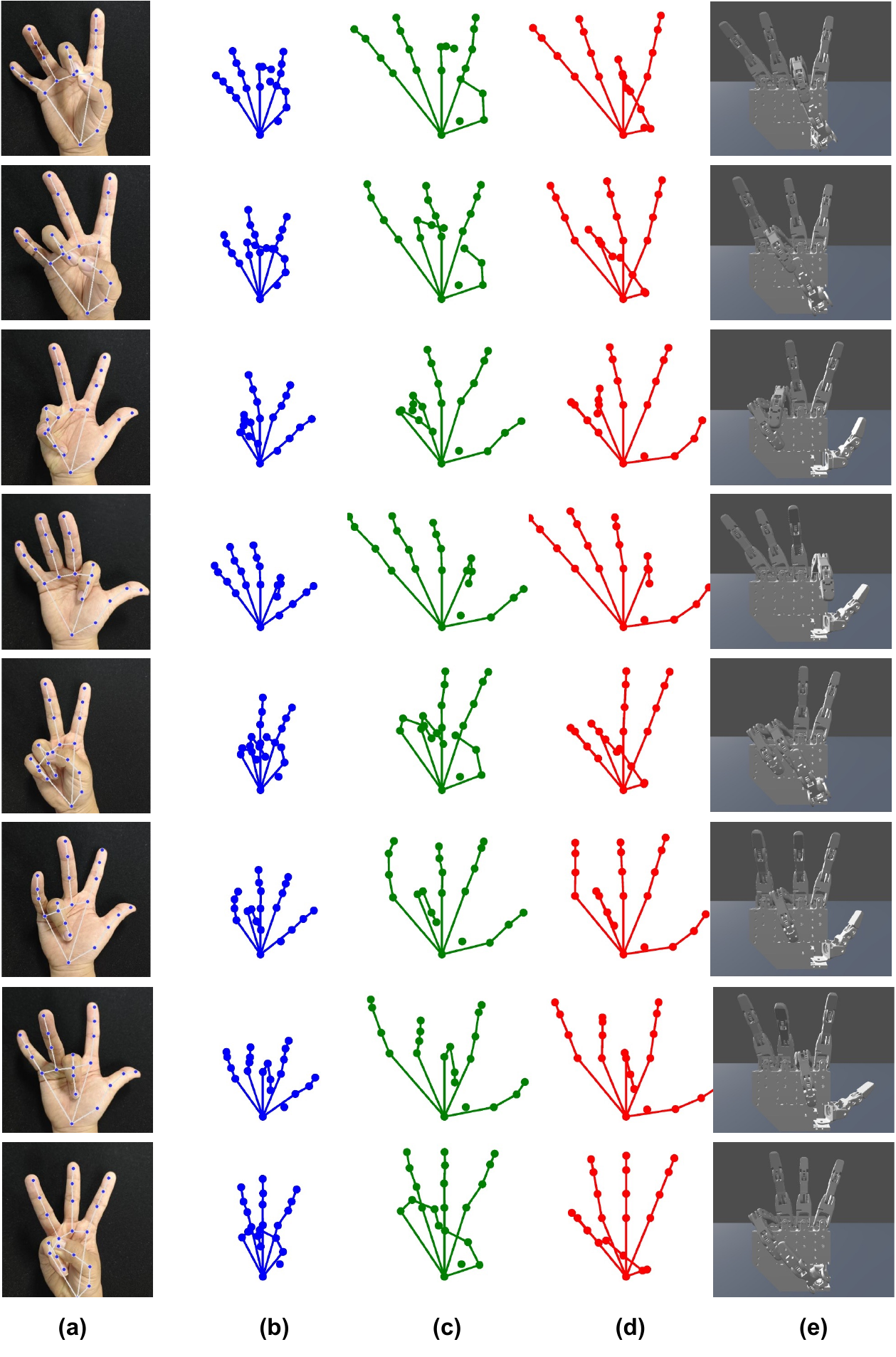}
\end{center}
\vspace{-10pt}
\caption{\textbf{Optimized Hand Retargeting Process}.  (a) Detected gesture with rocognized keypoints; (b) 3D keypoints corresponding to (a); (c) Refined gesture keypoints using ~\eqref{eq:gesture_keypoint_correction}; (d) Optimized robotic hand keypoints; (e) RAPID hand poses}
\label{fig: retargeting_2}
\vspace{-15pt}
\end{figure}

Among these systems, the MoCap system provides the best tracking performance in terms of precision and robustness, particularly within a 1-1.5m range. The Vision Pro strikes a balance between cost and performance but struggles with pinky finger tracking due to occlusion. The camera system is the most affordable but has the weakest performance. We also visualize the refined retargeting process across various hand poses using the proposed interface, as shown in Fig. \ref{fig: retargeting_2}.

\subsubsection{Real-world test} We further conduct real-world tests via an Apple Vision Pro across three challenging tasks: multi-finger retrieval, ladle handling, and human-like piano playing. These tasks pose significant challenges for robotic dexterous manipulation.

\textbf{Multi-finger retrieval}. 
Most object retrieval tasks \cite{zeng2018learning, tang2021learning, yang2021collaborative, xu2024geotact} rely on parallel grippers combined with arm motion, leading to non-prehensile pushing. In contrast, we design a more realistic scenario where the target object is buried in a densely stacked drawer, requiring multi-finger, non-prehensile manipulation to search for and retrieve the object. This task reflects daily challenges such as finding items in a cluttered drawer. As shown in Fig. \ref{fig: tasks} (a), the RAPID Hand completes this task due to its thin finger design, with motors and cables positioned in the palm instead of inside the fingers. This suggests that a non-bulky finger design should be necessary for non-prehensile dexterous manipulation.

\textbf{Ladle handling}. 
Unlike previous studies \cite{wang2024dexcap, lin2024learning, ding2024bunny, cheng2024open}, where tools are simplified or pre-positioned for easy grasping, this task involved picking up a ladle from a flat surface and adjusting its in-hand position using multi-finger manipulation. This mirrors a common human activity and presents a long-horizon challenge, requiring fine manipulation to adjust the ladle's orientation. As shown in Fig. \ref{fig: tasks} (b), the RAPID Hand is able to grasp and adjust the ladle, though occasional failures occur. These failures are primarily due to difficulty grasping the ladle's shank from the table, which requires precise fingertip control. This shows that fully actuated hands may offer advantages in in-hand manipulation, though further refinements would likely improve performance.

\textbf{Human-like Piano Playing}. While human-like piano playing has been explored in simulation \cite{robopianist2023, qian2024pianomime}, it remains a challenging task in real-world settings. In this task, the robot plays an electronic piano, using both arm movement and finger dexterity to mimic human performance. This is particularly difficult for the human operator due to partial observations. We demonstrate this task through teleoperation using the RAPID Hand. While the result may not be ideal, it highlights the significant potential of an affordable, fully actuated 20-DoF design in handling complex tasks and suggests that such designs are crucial for future advancements.

\section{CONCLUSION}
\label{sec: conclusion}

In this paper, we present the RAPID Hand, a low-cost, biomimetic dexterous hand system designed to enhance dexterous teleoperation tasks. We introduce a novel anthropomorphic actuation and transmission scheme, alongside an intuitive teleoperation interface optimized for high DoA hands. Experimental results demonstrate the value of an affordable 20-DoA design and the effectiveness of the retargeting method. However, there is still room for improvement. We hope that the open-source of our design will encourage further innovation and collaboration within the robotics community. 

\section{ACKNOWLEDGEMENT}
\label{sec: acknowledgement}
We would also like to thank LUSTER for motion capture devices. We sincerely appreciate the contributions from all related institutions and authors.

\label{sec:subjective}

{
\bibliographystyle{IEEEtran}
\bibliography{IEEEabrv,reference}

\begin{thebibliography}{10}
\providecommand{\url}[1]{#1}
\csname url@samestyle\endcsname
\providecommand{\newblock}{\relax}
\providecommand{\bibinfo}[2]{#2}
\providecommand{\BIBentrySTDinterwordspacing}{\spaceskip=0pt\relax}
\providecommand{\BIBentryALTinterwordstretchfactor}{4}
\providecommand{\BIBentryALTinterwordspacing}{\spaceskip=\fontdimen2\font plus
\BIBentryALTinterwordstretchfactor\fontdimen3\font minus \fontdimen4\font\relax}
\providecommand{\BIBforeignlanguage}[2]{{%
\expandafter\ifx\csname l@#1\endcsname\relax
\typeout{** WARNING: IEEEtran.bst: No hyphenation pattern has been}%
\typeout{** loaded for the language `#1'. Using the pattern for}%
\typeout{** the default language instead.}%
\else
\language=\csname l@#1\endcsname
\fi
#2}}
\providecommand{\BIBdecl}{\relax}
\BIBdecl

\bibitem{ma2011dexterity}
R.~R. Ma and A.~M. Dollar, ``On dexterity and dexterous manipulation,'' in \emph{2011 15th International Conference on Advanced Robotics (ICAR)}.\hskip 1em plus 0.5em minus 0.4em\relax IEEE, 2011, pp. 1--7.

\bibitem{andrychowicz2020learning}
O.~M. Andrychowicz, B.~Baker, M.~Chociej, R.~Jozefowicz, B.~McGrew, J.~Pachocki, A.~Petron, M.~Plappert, G.~Powell, A.~Ray \emph{et~al.}, ``Learning dexterous in-hand manipulation,'' \emph{The International Journal of Robotics Research}, vol.~39, no.~1, pp. 3--20, 2020.

\bibitem{chen2023sequential}
Y.~Chen, C.~Wang, L.~Fei-Fei, and C.~K. Liu, ``Sequential dexterity: Chaining dexterous policies for long-horizon manipulation,'' \emph{arXiv preprint arXiv:2309.00987}, 2023.

\bibitem{chi2023diffusionpolicy}
C.~Chi, S.~Feng, Y.~Du, Z.~Xu, E.~Cousineau, B.~Burchfiel, and S.~Song, ``Diffusion policy: Visuomotor policy learning via action diffusion,'' in \emph{Proceedings of Robotics: Science and Systems (RSS)}, 2023.

\bibitem{qi2023general}
H.~Qi, B.~Yi, S.~Suresh, M.~Lambeta, Y.~Ma, R.~Calandra, and J.~Malik, ``General in-hand object rotation with vision and touch,'' in \emph{Conference on Robot Learning}.\hskip 1em plus 0.5em minus 0.4em\relax PMLR, 2023, pp. 2549--2564.

\bibitem{zhao2023learning}
T.~Z. Zhao, V.~Kumar, S.~Levine, and C.~Finn, ``Learning fine-grained bimanual manipulation with low-cost hardware,'' \emph{arXiv preprint arXiv:2304.13705}, 2023.

\bibitem{fu2024mobile}
Z.~Fu, T.~Z. Zhao, and C.~Finn, ``Mobile aloha: Learning bimanual mobile manipulation with low-cost whole-body teleoperation,'' \emph{arXiv preprint arXiv:2401.02117}, 2024.

\bibitem{chi2024universal}
C.~Chi, Z.~Xu, C.~Pan, E.~Cousineau, B.~Burchfiel, S.~Feng, R.~Tedrake, and S.~Song, ``Universal manipulation interface: In-the-wild robot teaching without in-the-wild robots,'' in \emph{Proceedings of Robotics: Science and Systems (RSS)}, 2024.

\bibitem{wang2024dexcap}
C.~Wang, H.~Shi, W.~Wang, R.~Zhang, L.~Fei-Fei, and C.~K. Liu, ``Dexcap: Scalable and portable mocap data collection system for dexterous manipulation,'' \emph{arXiv preprint arXiv:2403.07788}, 2024.

\bibitem{fang2024airexo}
H.~Fang, H.-S. Fang, Y.~Wang, J.~Ren, J.~Chen, R.~Zhang, W.~Wang, and C.~Lu, ``Airexo: Low-cost exoskeletons for learning whole-arm manipulation in the wild,'' in \emph{2024 IEEE International Conference on Robotics and Automation (ICRA)}.\hskip 1em plus 0.5em minus 0.4em\relax IEEE, 2024, pp. 15\,031--15\,038.

\bibitem{allegrohand}
``Allegrohand,'' \url{https://www.wonikrobotics.com/research-robot-hand}, accessed: 2024-09-15.

\bibitem{shadowhand}
``Shadowhand,'' \url{https://ninjatek.com/shop/edge/}, accessed: 2024-09-15.

\bibitem{inspirehand}
``Inspirehand,'' \url{https://inspire-robots.store}, accessed: 2024-09-15.

\bibitem{shaw2023leap}
K.~Shaw, A.~Agarwal, and D.~Pathak, ``Leap hand: Low-cost, efficient, and anthropomorphic hand for robot learning,'' \emph{arXiv preprint arXiv:2309.06440}, 2023.

\bibitem{abilityhand}
``Abilityhand,'' \url{https://www.psyonic.io/ability-hand}, accessed: 2024-09-15.

\bibitem{vint6d}
Z.~Wan, Y.~Ling, S.~Yi, L.~Qi, M.~Lu, W.~Lee, S.~Yang, X.~Teng, P.~Lu, X.~Yang, M.-H. Yang, and H.~Cheng, ``Vint-6d: A large-scale object-in-hand dataset from vision, touch and proprioception,'' in \emph{International Conference on Machine Learning}, 2024.

\bibitem{barretthand}
``Barretthand,'' \url{https://ninjatek.com/shop/edge/}, accessed: 2024-09-15.

\bibitem{butterfass2001dlr}
J.~Butterfa{\ss}, M.~Grebenstein, H.~Liu, and G.~Hirzinger, ``Dlr-hand ii: Next generation of a dextrous robot hand,'' in \emph{Proceedings 2001 ICRA. IEEE International Conference on Robotics and Automation (Cat. No. 01CH37164)}, vol.~1.\hskip 1em plus 0.5em minus 0.4em\relax IEEE, 2001, pp. 109--114.

\bibitem{toshimitsu2023getting}
Y.~Toshimitsu, B.~Forrai, B.~G. Cangan, U.~Steger, M.~Knecht, S.~Weirich, and R.~K. Katzschmann, ``Getting the ball rolling: Learning a dexterous policy for a biomimetic tendon-driven hand with rolling contact joints,'' in \emph{2023 IEEE-RAS 22nd International Conference on Humanoid Robots (Humanoids)}.\hskip 1em plus 0.5em minus 0.4em\relax IEEE, 2023, pp. 1--7.

\bibitem{chalon2010thumb}
M.~Chalon, M.~Grebenstein, T.~Wimb{\"o}ck, and G.~Hirzinger, ``The thumb: Guidelines for a robotic design,'' in \emph{2010 IEEE/RSJ international conference on intelligent robots and systems}.\hskip 1em plus 0.5em minus 0.4em\relax IEEE, 2010, pp. 5886--5893.

\bibitem{yoshikawa1985manipulability}
T.~Yoshikawa, ``Manipulability of robotic mechanisms,'' \emph{The international journal of Robotics Research}, vol.~4, no.~2, pp. 3--9, 1985.

\bibitem{brohan2022rt}
A.~Brohan, N.~Brown, J.~Carbajal, Y.~Chebotar, J.~Dabis, C.~Finn \emph{et~al.}, ``Rt-1: Robotics transformer for real-world control at scale,'' \emph{Proceedings of Robotics: Science and Systems (RSS)}, 2023.

\bibitem{pmlr-v229-zitkovich23a}
B.~Zitkovich, T.~Yu \emph{et~al.}, ``Rt-2: Vision-language-action models transfer web knowledge to robotic control,'' in \emph{Proceedings of The 7th Conference on Robot Learning (CoRL)}, 2023, pp. 2165--2183.

\bibitem{RTX}
A.~O’Neill, A.~Rehman, A.~Maddukuri \emph{et~al.}, ``Open x-embodiment: Robotic learning datasets and rt-x models : Open x-embodiment collaboration,'' in \emph{2024 IEEE International Conference on Robotics and Automation (ICRA)}, 2024, pp. 6892--6903.

\bibitem{ding2024bunny}
R.~Ding, Y.~Qin, J.~Zhu, C.~Jia, S.~Yang, R.~Yang, X.~Qi, and X.~Wang, ``Bunny-visionpro: Real-time bimanual dexterous teleoperation for imitation learning,'' \emph{arXiv preprint arXiv:2407.03162}, 2024.

\bibitem{lin2024learning}
T.~Lin, Y.~Zhang, Q.~Li, H.~Qi, B.~Yi, S.~Levine, and J.~Malik, ``Learning visuotactile skills with two multifingered hands,'' \emph{arXiv preprint arXiv:2404.16823}, 2024.

\bibitem{cerulo2017teleoperation}
I.~Cerulo, F.~Ficuciello, V.~Lippiello, and B.~Siciliano, ``Teleoperation of the schunk s5fh under-actuated anthropomorphic hand using human hand motion tracking,'' \emph{Robotics and Autonomous Systems}, vol.~89, pp. 75--84, 2017.

\bibitem{liu2019high}
H.~Liu, Z.~Zhang, X.~Xie, Y.~Zhu, Y.~Liu, Y.~Wang, and S.-C. Zhu, ``High-fidelity grasping in virtual reality using a glove-based system,'' in \emph{2019 international conference on robotics and automation (icra)}.\hskip 1em plus 0.5em minus 0.4em\relax IEEE, 2019, pp. 5180--5186.

\bibitem{qin2023anyteleop}
Y.~Qin, W.~Yang, B.~Huang, K.~Van~Wyk, H.~Su, X.~Wang, Y.-W. Chao, and D.~Fox, ``Anyteleop: A general vision-based dexterous robot arm-hand teleoperation system,'' \emph{arXiv preprint arXiv:2307.04577}, 2023.

\bibitem{fu2024humanplus}
Z.~Fu, Q.~Zhao, Q.~Wu, G.~Wetzstein, and C.~Finn, ``Humanplus: Humanoid shadowing and imitation from humans,'' \emph{arXiv preprint arXiv:2406.10454}, 2024.

\bibitem{sivakumar2022robotic}
A.~Sivakumar, K.~Shaw, and D.~Pathak, ``Robotic telekinesis: Learning a robotic hand imitator by watching humans on youtube,'' \emph{arXiv preprint arXiv:2202.10448}, 2022.

\bibitem{kim2024surgical}
J.~W. Kim, T.~Z. Zhao, S.~Schmidgall, A.~Deguet, M.~Kobilarov, C.~Finn, and A.~Krieger, ``Surgical robot transformer (srt): Imitation learning for surgical tasks,'' \emph{arXiv preprint arXiv:2407.12998}, 2024.

\bibitem{xu2025robopanoptes}
X.~Xu, D.~Bauer, and S.~Song, ``Robopanoptes: The all-seeing robot with whole-body dexterity,'' \emph{arXiv preprint arXiv:2501.05420}, 2025.

\bibitem{wang2024lessons}
J.~Wang, Y.~Yuan, H.~Che, H.~Qi, Y.~Ma, J.~Malik, and X.~Wang, ``Lessons from learning to spin" pens",'' \emph{arXiv preprint arXiv:2407.18902}, 2024.

\bibitem{lin2024twisting}
T.~Lin, Z.-H. Yin, H.~Qi, P.~Abbeel, and J.~Malik, ``Twisting lids off with two hands,'' \emph{arXiv preprint arXiv:2403.02338}, 2024.

\bibitem{yin2024learning}
J.~Yin, H.~Qi, J.~Malik, J.~Pikul, M.~Yim, and T.~Hellebrekers, ``Learning in-hand translation using tactile skin with shear and normal force sensing,'' \emph{arXiv preprint arXiv:2407.07885}, 2024.

\bibitem{shawleap}
K.~Shaw and D.~Pathak, ``Leap hand v2: Dexterous, low-cost anthropomorphic hybrid rigid soft hand for robot learning,'' in \emph{2nd Workshop on Dexterous Manipulation: Design, Perception and Control (RSS)}, 2024.

\bibitem{lee2016kitech}
D.-H. Lee, J.-H. Park, S.-W. Park, M.-H. Baeg, and J.-H. Bae, ``Kitech-hand: A highly dexterous and modularized robotic hand,'' \emph{IEEE/ASME Transactions on Mechatronics}, vol.~22, no.~2, pp. 876--887, 2016.

\bibitem{boggs1995sequential}
P.~T. Boggs and J.~W. Tolle, ``Sequential quadratic programming,'' \emph{Acta numerica}, vol.~4, pp. 1--51, 1995.

\bibitem{johnson2021nlopt}
S.~G. Johnson and J.~Schueller, ``Nlopt: Nonlinear optimization library,'' \emph{Astrophysics Source Code Library}, pp. ascl--2111, 2021.

\bibitem{Xiang_2020_SAPIEN}
F.~Xiang, Y.~Qin, K.~Mo, Y.~Xia, H.~Zhu, F.~Liu, M.~Liu, H.~Jiang, Y.~Yuan, H.~Wang, L.~Yi, A.~X. Chang, L.~J. Guibas, and H.~Su, ``{SAPIEN}: A simulated part-based interactive environment,'' in \emph{The IEEE Conference on Computer Vision and Pattern Recognition (CVPR)}, June 2020.

\bibitem{zeng2018learning}
A.~Zeng, S.~Song, S.~Welker, J.~Lee, A.~Rodriguez, and T.~Funkhouser, ``Learning synergies between pushing and grasping with self-supervised deep reinforcement learning,'' in \emph{2018 IEEE/RSJ International Conference on Intelligent Robots and Systems (IROS)}.\hskip 1em plus 0.5em minus 0.4em\relax IEEE, 2018, pp. 4238--4245.

\bibitem{tang2021learning}
B.~Tang, M.~Corsaro, G.~Konidaris, S.~Nikolaidis, and S.~Tellex, ``Learning collaborative pushing and grasping policies in dense clutter,'' in \emph{2021 IEEE International Conference on Robotics and Automation (ICRA)}.\hskip 1em plus 0.5em minus 0.4em\relax IEEE, 2021, pp. 6177--6184.

\bibitem{yang2021collaborative}
Y.~Yang, Z.~Ni, M.~Gao, J.~Zhang, and D.~Tao, ``Collaborative pushing and grasping of tightly stacked objects via deep reinforcement learning,'' \emph{IEEE/CAA Journal of Automatica Sinica}, vol.~9, no.~1, pp. 135--145, 2021.

\bibitem{xu2024geotact}
J.~Xu, Y.~Jia, D.~Yang, P.~Meng, X.~Zhu, Z.~Guo, S.~Song, and M.~Ciocarlie, ``Tactile-based object retrieval from granular media,'' \emph{arXiv preprint arXiv:2402.04536}, 2023.

\bibitem{cheng2024open}
X.~Cheng, J.~Li, S.~Yang, G.~Yang, and X.~Wang, ``Open-television: Teleoperation with immersive active visual feedback,'' \emph{arXiv preprint arXiv:2407.01512}, 2024.

\bibitem{robopianist2023}
K.~Zakka, P.~Wu, L.~Smith, N.~Gileadi, T.~Howell, X.~B. Peng, S.~Singh, Y.~Tassa, P.~Florence, A.~Zeng, and P.~Abbeel, ``Robopianist: Dexterous piano playing with deep reinforcement learning,'' in \emph{Conference on Robot Learning (CoRL)}, 2023.

\bibitem{qian2024pianomime}
C.~Qian, J.~Urain, K.~Zakka, and J.~Peters, ``Pianomime: Learning a generalist, dexterous piano player from internet demonstrations,'' \emph{arXiv preprint arXiv:2407.18178}, 2024.

\end{thebibliography}
}

\end{document}